\documentclass[conference]{IEEEtran}
\IEEEoverridecommandlockouts
\usepackage{cite}
\usepackage{amsmath,amssymb,amsfonts}
\usepackage{algorithmic}
\usepackage{graphicx}
\usepackage{textcomp}
\def\BibTeX{{\rm B\kern-.05em{\sc i\kern-.025em b}\kern-.08em
    T\kern-.1667em\lower.7ex\hbox{E}\kern-.125emX}}

\usepackage{multirow}
\newcommand{\eat}[1]{}

\begin{document}
\title{Unsupervised Adversarial Attacks on Deep Feature-based Retrieval with GAN  \\
\thanks{\IEEEauthorrefmark{1} Corresponding Author}
}

\author{\IEEEauthorblockN{Guoping Zhao\IEEEauthorrefmark{0},
Mingyu Zhang\IEEEauthorrefmark{0},
Jiajun Liu\IEEEauthorrefmark{1}, and
Ji-Rong Wen\IEEEauthorrefmark{0}}
\IEEEauthorblockA{\IEEEauthorrefmark{0}Beijing Key Laboratory of Big Data Management and Analysis Methods\\
School of Information, Renmin University of China\\}
}

\maketitle

\begin{abstract}
Studies show that Deep Neural Network (DNN)-based image classification models are vulnerable to maliciously constructed adversarial examples. 
  However, little effort has been made to investigate how DNN-based image retrieval models are affected by such attacks. 
  
  In this paper, we introduce Unsupervised Adversarial Attacks with Generative Adversarial Networks (UAA-GAN) to attack deep feature-based image retrieval systems.
  UAA-GAN is an unsupervised learning model that requires only a small amount of unlabeled data for training. 
  Once trained, it produces query-specific perturbations for query images to form adversarial queries. 
  The core idea is to ensure that the attached perturbation is barely perceptible to human yet effective in pushing the query away from its original position in the deep feature space. 
  
  UAA-GAN works with various application scenarios that are based on deep features, including image retrieval, person Re-ID and face search. 
  Empirical results show that UAA-GAN cripples retrieval performance without significant visual changes in the query images. UAA-GAN generated adversarial examples are less distinguishable because they tend to incorporate subtle perturbations in textured or salient areas of the images, such as key body parts of human, dominant structural patterns/textures or edges, rather than in visually insignificant areas (e.g., background and sky). 
  Such tendency indicates that the model indeed learned how to toy with both image retrieval systems and human eyes.\end{abstract}


\section{Introduction}
\begin{figure}[htbp]
  \centering
  \includegraphics[width=8.5cm]{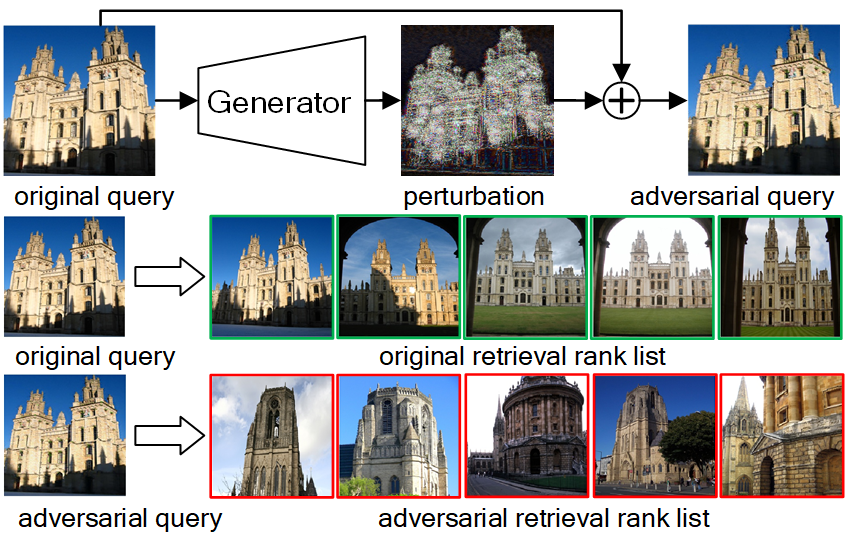}
  \caption{
    An example of adversarial attack on deep feature-based image retrieval systems. 
    It is evident that although visual difference between the adversarial example (top right and bottom left) and the original image (top left and middle left) is very small, the adversarial query image can fool the retrieval system to return visually dissimilar search results.
    Green borders of the results mean correct query results and red borders mean otherwise.
    }
  \label{fig1}
\end{figure}

Deep neural networks (DNNs) are a powerful feature representation learning tool that achieves state-of-the-art performance in content-based image retrieval (CBIR).
In recent years, deep features are quickly replacing conventional image features that rely on handcrafted key-point detectors and descriptors. 
DNN-based models generate a deep feature descriptor for an image by aggregating the activations from the top layers of a pre-trained deep neural network, and then the similarity(or distance) between two images is determined by the euclidean distance or cosine similarity of their feature vectors. Such methods have been observed to be able to preserve more abstract and global semantic information than those low-level key-point-based features.
Inspired by the excellent representation capability of DNN, many researchers focus on improving the retrieval accuracy through learning a discriminative feature representation.
However, the robustness and stability of DNN features in the retrieval task has been largely overlooked.
As shown in Figure ~\ref{fig1}, a DNNs-based image retrieval system is successfully fooled by a generated adversarial query image, which is visually nearly identical to the original copy yet has a very different deep feature representation .
\vspace{-1em}

It is known that DNNs-based classification systems are vulnerable to adversarial examples: by adding some carefully crafted, minor perturbations to an input image, the target DNN is often mislead and predicts the image to a wrong class with high confidence \cite{Szegedy2014ICLR}.
Adversarial examples are first introduced in \cite{Szegedy2014ICLR}, the perturbations are generated by tweaking the input for maximized  classification error.
Much attention has then been drawn to the rivalry between adversarial attacks and defenses.
The Fast Gradient Sign Method (FSGM) \cite{Ian2015iclr} and its iterative variation \cite{kurakin2016adversarial} generate adversarial examples through gradient updates along the direction that maximizes the  loss.
Optimization-based methods, such as C\&W \cite{carlini2017towards} \eat{ and ZOO \cite{chen2017zoo}}, reformulate adversarial attacks on image classification as an optimization problem that generates more sophisticated adversarial examples.
Leveraging the idea from DeepFool \cite{moosavi2016deepfool}, Moosavi-Dezfooli et al. \cite{moosavi2017universal} propose a method image-agnostic adversarial perturbations, named Universal Adversarial Perturbations (UAP), which makes natural images getting misclassified by the target network.

More recently, researchers started to examine the effect of adversarial examples on other tasks, such as object detection \cite{xie2017adversarial}, semantic segmentation \cite{xie2017adversarial, Metzen_2017_ICCV}, image caption \cite{acl_chen_2018} and face recognition \cite{Mahmood2016accessorize}.
Despite the increasing research attention on adversarial examples, image retrieval as a task has yet to be studied thoroughly as a target for such attacks.
In this work, we demonstrate that deep feature-based Content Based Image Retrieval (CBIR) systems and its derived applications: person re-identification (ReID) and face search, are also prone to adversarial attacks. That is, it is possible to tamper with an image to make it almost visually identical to its original form yet nearly impossible to be searched in an image retrieval system.
Attacking image retrieval system is significantly more challenging than attacking classification models, due to two main reasons.

First reason is, in image classification, the output of the fully-connected layer and the softmax activation function are very sensitive to minor changes around key points and areas on the target object to be classified. Minor perturbations can significantly reduce the activation of the pre-trained convolution kernels that are supposed to function as local feature detectors.
    Whereas on the image retrieval task, the image is searched against a database on global features that went through many layers of max-pooling or sum-pooling layers of convolutional neural networks (CNNs), hence it is in general more invariant and robust to minor local changes.
    
    Then, class labels are normally available for training the classifier, the adversarial example generation is hence easy to formulate. 
    By using the gradient information in the classification process, it is quite intuitive to generate adversarial examples by pushing the input away across the decision boundary.
    While for image retrieval, the goal is to push the feature representation away from its original position and its original neighbors in the feature space.
    This problem is more difficult to formulate because it lacks of well defined labels and gradient information.

We propose a novel Unsupervised Adversarial Attacks method with GAN (UAA-GAN) that aims at fooling deep feature-based image retrieval systems, by giving it three desirable properties: 
Firstly, the generated adversarial example should be distant from the original image and its original neighbors;
Secondly, the norm and placement of the perturbation should be in such a way that the visual difference of the adversarial example from the original image is not easily identifiable.
Thirdly, the adversarial examples should realistic-looking and high perceptual quality.

Our main contributions are threefold:
\begin{itemize}
  \item We proposed an efficient GAN-based attack framework called UAA-GAN for generating photo realistic adversarial examples with effective perturbations nearly imperceptibility to humans.
  \item UAA-GAN is entire unsupervised and requires only a small amount of unlabeled images for training.
        Once trained, it can generate perturbations according to each query image.
  \item We evaluate UAA-GAN on three tasks: content-based image retrieval, person re-identification and face search.
  Empirical results show high effectiveness of UAA-GAN in all tasks. 
\end{itemize}

The rest of the paper proceeds as follows:
Section \ref{method} describes the technical details of the proposed UAA-GAN framework. 
Section \ref{experiment} reports comprehensive experimental results and analysis.
The relevant existing works are described in Section \ref{related} . 
Finally, we conclude our work in section \ref{conclusion}.

\begin{figure*}[htb]
  \centering
  \includegraphics[width=17.5cm]{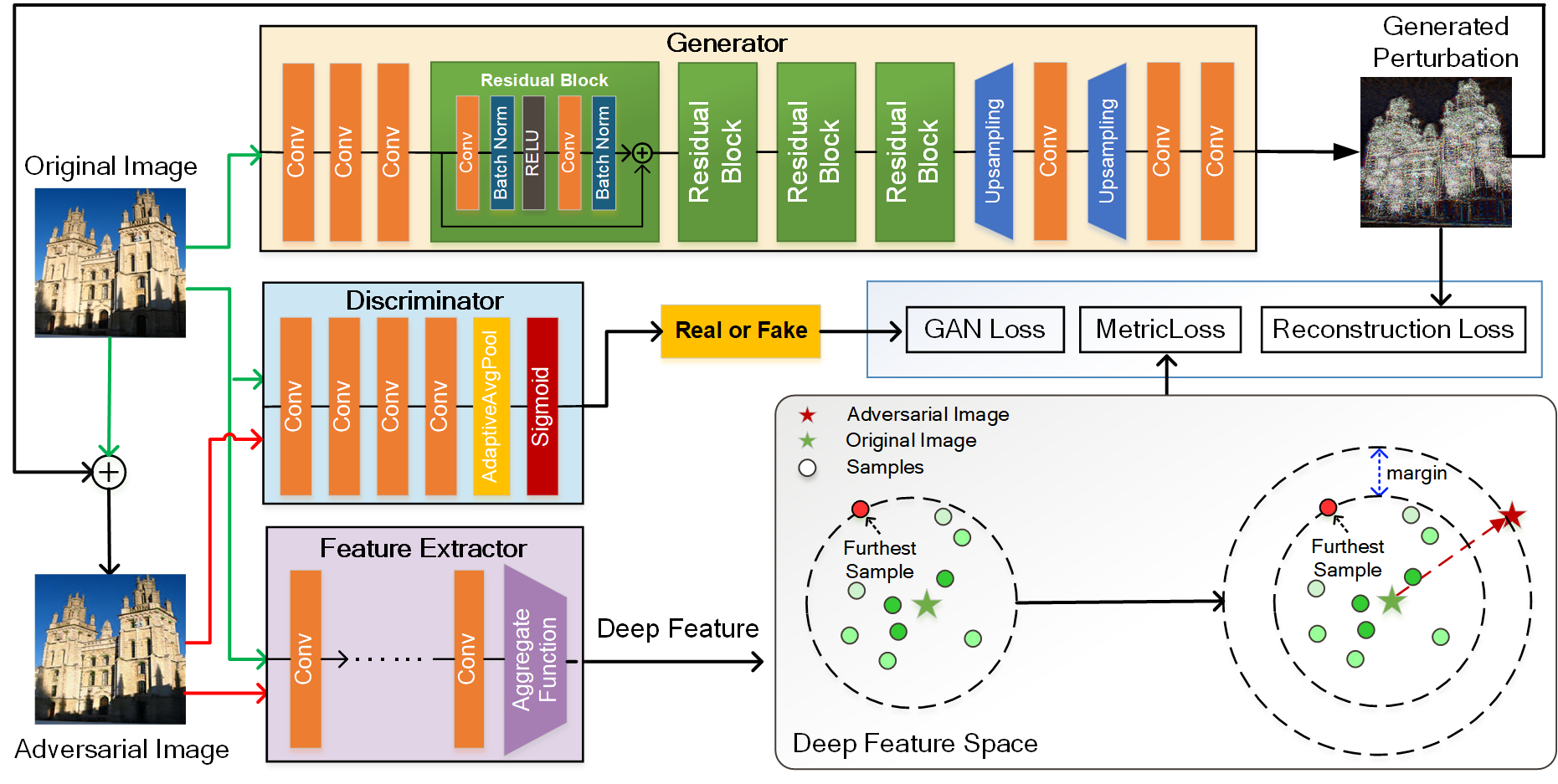}
  \centering
  \caption{The overall architecture of the UAA-GAN framework. 
    The design goal of this framework is to attack a retrieval system built on a specific target feature extraction network by generating an adversarial example with query-specific perturbation. By considering three different losses, we ensure three objectives are achieved with the adversarial example.  The GAN loss enforces that the adversarial example, changed by perturbation though, should look as natural as possible; the metric loss is used to push the adversarial example aways from the original query's feature in the deep feature space; the reconstruction loss acts as a regularization to restrict the level of perturbation that is added to the original query image.} 
  \label{fig2}
\end{figure*}

\section{Related Work}\label{related}

\subsection{Adversarial Examples}
The adversarial example method is proposed in \cite{Szegedy2014ICLR}, which proves small and intentional perturbations can mislead machine learning models to make false predictions.
After that, many methods have been devised to construct adversarial examples.
Goodfellow et al. \cite{Ian2015iclr} proposed a simple fast gradient sign method (FSGM), which only adds a small error multiplied by the sign of the gradients to the input to form an adversarial example.
Basic Iterative Method (BIM) \cite{kurakin2016adversarial} is the iterative version of FSGM, which produces better adversarial images through applying gradient update and clipping repetitively.
AdvGAN \cite{Xiao2018ijcai} applies generative adversarial networks to produce adversarial examples.
Unlike methods that generate different perturbations for each image, Universal Adversarial Perturbations (UAP) \cite{moosavi2017universal} computes an image-agnostic and minor perturbations to fool deep networks.
Recently, adversarial attacks have been extended to many other applications.
Xie et al. \cite{xie2017adversarial} introduce a novel Dense Adversary Generation (DAG) algorithm to fool the networks for semantic segmentation and object detection.
The Show-and-Fool method presented in \cite{acl_chen_2018} suggests that neural image caption models can also be vulnerable to attacks.

\subsection{Deep Feature-based Image Retrieval}
Image retrieval aims to search the image database for similar images as the query.
It has been widely used in many applications, such as image search engines or product search engines such as that on eBay \cite{yang2017visual}.
Latest DNN-based retrieval methods represent images by aggregated deep features extracted from pre-trained or fine-tuned CNN models. 
The similarity between two images is directly measured by the euclidean distance or cosine similarity of two image representations in the deep feature space.
The simplest deep global descriptors aggregate compact features by sum-pooling (SPoC \cite{Babenko_2015_ICCV}) or max-pooling (MAC \cite{tolias2015particular}) the feature maps output by convolutional layers.
Regional Maximum Activation of Convolutions (R-MAC) \cite{tolias2015particular} extends the MAC by summing the MAC features of several regions at multiple scales.
There are methods that further improve the performance of image retrieval by fine-tuning the backbone networks (such as: VGG, ResNet) on task-related datasets.
NetVLAD \cite{arandjelovic2016netvlad}, which adds a VLAD pooling layer after convolutional layers, is trained in an end-to-end manner by a weakly supervised ranking loss.
The training data of NetVLAD is collected from the Google Street View Time Machine.
Filip et al. proposed a trainable Generalized-Mean (GeM) pooling layer in \cite{radenovic2018fine}, and fine-tunes the network using a dataset collected from Flickr through structure-from-motion (SfM) 3D reconstruction algorithm\eat{\cite{schonberger2015single}}.
DELF \cite{noh2017large}\eat{ and MSCAN \cite{lou2018multi}} introduces attention mechanism to the base networks to obtain discriminative image representations.

Person ReID and face search are the two most essential derived subproblems of image retrieval.
They aim at searching in the gallery for images containing the same person/face with the query image. 
In recent years, most of the state-of-the-art methods are based on DNNs.
We choose the state-of-the-art methods to evaluate the attack performance of UAA-GAN, such as MGN \cite{wang2018learning} for ReID task and Sphereface \cite{liu2017sphereface} for face search task.

\subsection{Generative Adversarial Networks}
Generative Adversarial Networks (GANs) are first proposed by Goodfellow et al. \cite{Ian2014NIPS}, who formulate the GAN as a two-player minimax game between two adversarial networks. 
Radford et al. proposed Deep convolutional GANs (DCGANs) \cite{Radford201515DCGAN}, which introduced convolutional layers and convolutional-transpose layers to GANs architecture. 
Conditional GANs (cGANs) \cite{mirza2014conditional} extend the vanilla GANs with conditional settings, and are able to control the generated samples with a condition via embedding the condition vector with the noise vector input to the generator.
GAN-based methods have achieved excellent performance on many image-to-image translation tasks, like image super resolution \cite{ledig2017photo}, image deblurring \cite{Kupyn_2018_CVPR},\eat{ style transfer \cite{wei2018person} ,} image synthesis \cite{huang2017beyond}, etc.
Despite the tremendous successes, GANs still suffer from the challenges of model collapse and instability in training.
Many studies have been proposed to address these problems by improving the optimization objectives.
WGANs \cite{arjovsky2017wasserstein} improved the vanilla GANs by replacing the Jensen-Shannon divergence with the Wassertein distance.
WGAN-GP \cite{gulrajani2017improved} is an improved version of WGAN, which uses gradient penalty instead of the weight clipping to enforce the Lipschitz constraint.
LSGANs \cite{mao2017least} used the least squares loss function instead of the sigmoid cross entropy loss function for the discriminator, making the training process more stable. 
Xiao et al. \cite{Xiao2018ijcai} first introduced the GANs into the problem of adversarial examples.
They proposed AdvGAN to generate adversarial examples, in both semi-whitebox and black-box attack settings, on several different image classification models. 
Perceptual-Sensitive GAN (PS-GAN) \cite{Liu2019aaai} focus on attacking image classification models through generating a adversarial patch with GAN.
\eat{Zhao et al. \cite{zhao2018generating} utilized GANs as a sub-model of their framework to generate adversarial examples of images and texts.}

\section{Methodology}\label{method}

In general, UAA-GAN is inspired by the GAN framework, which in many studies prove effective to mislead image classification systems and make them return incorrect classification results. 
Also inspired by the method of residual learning \cite{He_2016_CVPR}, we let the generator learn the mapping from the real images to perturbations, and use a skip connection to add the perturbations to the real images.

\subsection{Problem Formulation}\label{formulation}
Given an input image $\boldsymbol{x}$ of size $H \times W$, the feature maps (or activations) from a convolutional layer $l$ are denote as $\chi \in \mathbb{R}^{c \times h \times w}$.
Let $T_{\theta}()$ be the target image retrieval network with parameters $\theta$ , and $F$ be the feature aggregate function (pooling function),  we denote $f$ as the global deep feature of $x$, and $f_{\boldsymbol{x}} = F( T_{\theta}( \boldsymbol{x} ) )$.
The similarity of two image ($\boldsymbol{x_i}$ and $\boldsymbol{x_j}$) is measured by calculating the distance between the deep features of the two images via a metric function $d(f_{\boldsymbol{x_i}},f_{\boldsymbol{x_j}})$.

We use $\delta$ to represent the perturbation generated by generator $G$ and $\boldsymbol{\tilde{x}}$ represents the adversarial example (with normalized color values), which can be formulated as: 
\begin{eqnarray}
  &&\delta = clip(G(\boldsymbol{x}), -\epsilon, \epsilon )        \label{eq1} \\ 
  &&\boldsymbol{\tilde{x}} = clip(\boldsymbol{x} + \delta, 0, 1)  \label{eq2}
\end{eqnarray}
where function $clip(input, min, max)$  is used to limit all elements in $input$ into the range $[min, max]$.
Here, we use $clip()$ to restrict the range of both the perturbations and the adversarial examples.

Deep image retrieval systems improve retrieval performance by three ways:
devising better aggregate functions (e.g., MAC \cite{tolias2015particular}, SPoC \cite{Babenko_2015_ICCV}, RMAC \cite{tolias2015particular}, GeM \cite{radenovic2018fine});
fine-tuning pre-trained networks under a more efficient objective function (e.g. classification loss \cite{noh2017large}, contrastive loss \cite{radenovic2018fine}, triplet loss \cite{gordo2017end});
using networks architectures that yield more distinctive features(e.g., attention networks \cite{noh2017large}).
In all three cases, the common purpose is to make similar images to have small distances in the deep features space, and vice versa. A successful attack method needs to invalidate such properties while keeping the adversarial example as real and similar to its original as possible.

The generated perturbation is expected to push the the query image away from the original image in the deep feature space. 
This can be described the following objective function:
\begin{eqnarray}
    maximize && d(f_{\boldsymbol{x}},f_{\boldsymbol{\tilde{x}}}) \nonumber \\ 
    s.t.  &&\left \| \delta \right \|_\infty \leq \epsilon , \\
         &&\boldsymbol{\tilde{x}} \in [0,1]  \nonumber
         \label{eqn1}
        \end{eqnarray}
However, if we only optimize Equation \ref{eqn1}, we will obtain an adversarial example that is visually and significantly different from the original.
This is because Equation \ref{eqn1} only limits the numeric range of disturbances, not the visual characteristics of the perturbations.
As GANs have achieved remarkable results in many image generation tasks recently, in this paper, we employ GANs to generate effective yet subtle perturbations.

\subsection{The Architecture of UAA-GAN}\label{architecture}
As depicted in Fig. ~\ref{fig2},  UAA-GAN consists of three components.
First, a generator that learns to generate effective yet subtle perturbations and subsequently produce adversarial query images.
Second, a discriminator that distinguishes generated adversarial examples from real images.
Third, a target network that computes deep feature of any input image.

\subsubsection{Generator} 
In UAA-GAN, the generator $G$ is used to generate query-specific perturbations by mapping the input query images to the adversarial perturbations manifold.
Unlike Gaussian noise, UAA-GAN-generated perturbations are expected to be image-specific and unevenly distributed to achieve maximized disturbance to retrieval results given a fixed level of noise.
UAA-GAN uses conditional GANs \cite{mirza2014conditional} as its generator backbone, which generates fake samples with given conditions.
The dimensions of the perturbation are identical as the input image, so we can build the generator using the encoder-decoder CNN architecture. 
G begins with a stride-1 convolution block and two stride-2 convolution blocks to down-sample and encode the input.
The convolution block for the down-sampling process consists of a convolutional layer, an instance normalization layer and a Rectified Linear Unit (ReLU) activation.
The size of kernels is 3 $\times$ 3 and the numbers of kernels are 8, 16, 32, respectively. 
Then we stack four residual blocks with skip-connections \cite{he2016identity} to further encode the features into a latent-space representation.
As show in Figure. \ref{fig2}, the residual blocks is composed of two stride-1 convolutional layers followed by a batch normalization layer, and a ReLU activation layer is connected after first convolutional layer.
Each convolutional layer in residual block contains 32 kernels with size 3 $\times$ 3.
The skip-connections in the residual blocks accelerate the convergence process by elevating the vanishing gradient phenomenon.
On the contrary, the decoder is a combination of two up-sampling blocks and a stride-1 convolution block.
As the deconvolution layer often creates checkerboard pattern of artifacts and results in blurred images \cite{odena2016deconvolution}, to obtain more realistic images, we use the resize-convolution approaches instead of deconvolution layer for up-sampling.
The size of convolution kernels in decoder is 3 $\times$ 3 and the numbers of kernels are 16, 8, 3, respectively.
The latent-space representation is sent to the decoder and we obtain the final perturbation, which is of the same dimensions as the input image.

\subsubsection{Discriminator} 
The role of the discriminator $D$ is to determine whether the input image is a generated adversarial example or a real image.
The architecture of $D$ is fairly straightforward: it consists of three convolution blocks, each of which is composed of a convolutional layer, a batch normalization layer, and a Leaky ReLU function.
For the convolutional layers in convolution block, we set the kernel size to 4 $\times$ 4, stride to 2, and the numbers of kernels are 8, 16 and 32, respectively.
The last layer of $D$ is a one kernel convolutional layer followed by a global average pooling layer and sigmoid function, which generates a 1-dimensional output.

\subsubsection{Target Network}
Regardless of the architecture of the feature extraction network, most deep feature-based image retrieval systems represent each image as a fixed-length feature vector.
The similarity between images is measured by a distance metric function, such as L2 distance or cosine similarity.
For UAA-GAN, details about the target network's architecture is insignificant, while the only information required is the distance metric for image features and the output feature vectors themselves. The architecture of the network could remain a black box to UAA-GAN.
Therefore UAA-GAN provides a means of attack with excellent transferability and generalizability.

\subsection{Loss Function}\label{loss}
A well-designed adversarial example has to meet two criterion: effective attack results and minimally-perceptible visual changes.
\eat{For UAA-GAN, the optimization target is to make the perturbations as inconspicuous as possible, while significantly reducing the  precision of the retrieval results.}
In order to achieve the above objectives, we incoporate three objective functions to jointly guide the learning process of the UAA-GAN.
They are the reconstruction loss, the GAN Loss and the metric loss.

\subsubsection{Reconstruction Loss}
The reconstruction loss is the Root-Mean-Squared Error (RMSE) between the output and the input, which penalizes the generator for introducing  differences from the input to the output.

\eat{MSE measures the average squared error between the adversarial example and the original image.}
We minimize the reconstruction loss to ensure that the adversarial example is as visually similar as possible to the original image.
The reconstruction loss is formulated as below:
\begin{eqnarray}
  \mathcal L_{recon} = \left \| \boldsymbol{\tilde{x}} - \boldsymbol{x} \right \|_2
\end{eqnarray}
Here $ \boldsymbol{x}$ is the original query image and $\boldsymbol{\tilde{x}}$ is the generated adversarial example.

\subsubsection{GAN Loss}
To ensure the generated examples look natural to human, we introduce the GAN loss to train the generator and discriminator.
In UAA-GAN, the goal of the generator is produce photo-realistic adversarial query images, so that the discriminator mistakenly considers it as a sample from the distribution of real images.
Meanwhile, the discriminator tries to classify the input image correctly into real or fake images.
For a better controlled training process, we replace the original cross entropy loss with the least squares loss  \cite{mao2017least} and.
The GAN loss functions for the generator and the discriminator can be defined as follows:
\begin{eqnarray}
  \mathcal L_{GAN\_D} = \mathbb{E}_{\boldsymbol{x} \sim p_{data(\boldsymbol{x})}}[(D(x) -1)^2] + \mathbb{E}_{\boldsymbol{\tilde{x}} \sim p_{( \boldsymbol{\tilde{x}} | \boldsymbol{x} )}}[(D(\boldsymbol{\tilde{x}}))^2]
\end{eqnarray}
\begin{eqnarray}
  && \mathcal L_{GAN\_G} = \mathbb{E}_{\boldsymbol{\tilde{x}} \sim p_{( \boldsymbol{\tilde{x}} | \boldsymbol{x} )} }[(D(\boldsymbol{\tilde{x}}) -1)^2]
\end{eqnarray}
where $p_{data(\boldsymbol{x})}$ is the distribution of real images and $p_{( \boldsymbol{\tilde{x}} | \boldsymbol{x} )}$ is the conditional distribution of adversarial examples given $\boldsymbol{x} \sim p_{data(\boldsymbol{x})}$.

\subsubsection{Metric Loss}
In UAA-GAN, the goal of metric loss is to push the adversarial example away from the original image and its neighbors in the deep feature space.
However, this simple intuition requires much effort to formulate into feasible optimziation objectives.
A large distance may lead to a steep increase in reconstruction loss or GAN loss, resulting in low quality generation results.
Conversely, a small distance may yield unsatisfactory attack results.
Therefore, we propose an adaptive strategy based on triplet loss \cite{schroff2015facenet} and online hard negative mining \cite{wang2014learning}, as illustrated at the bottom right of Figure \ref{fig2}.
 Let $<\boldsymbol{x}, \boldsymbol{\tilde{x}}, \boldsymbol{{x}'}>$ denote a triplet, where $\boldsymbol{x}$ is the original query image, $\boldsymbol{\tilde{x}}$ is the generated adversarial query image and $\boldsymbol{{x}'}$ is the hardest example (i.e. the real neighbor of $\boldsymbol{x}$ with the largest distance from $\boldsymbol{x}$ in the batch).
We aim to make the distance between $\boldsymbol{x}$ and $\boldsymbol{\tilde{x}}$ greater than that of $\boldsymbol{x}$ and $\boldsymbol{{x}'}$ by a given margin $m$.
The constraint can be written as:
\begin{eqnarray}
  d(f_{\boldsymbol{x}},f_{\boldsymbol{{x}'}}) + m+ \leq d(f_{\boldsymbol{x}},f_{\boldsymbol{\tilde{x}}})
\end{eqnarray}
where $m$ is a given scalar, used to control the margin.
The metric loss function is then defined as:
\begin{eqnarray}
  \mathcal{L}_{metric} =  max( d(f_{\boldsymbol{x}},f_{\boldsymbol{{x}'}}) + m - d(f_{\boldsymbol{x}},f_{\boldsymbol{\tilde{x}}}), 0 )  
\end{eqnarray}

Finally, the complete loss function used for training the UAA-GAN is the combination of the three components aforementioned:
\begin{eqnarray}
  && \mathcal L_{G} =  \mathcal L_{GAN\_G} + \lambda_r \mathcal L_{recon} + \lambda_m \mathcal{L}_{metric}  \label{eq9} \\
  && \mathcal L_{D} = \mathcal L_{GAN\_D} 
\end{eqnarray}
where $\lambda_r $ and $\lambda_m$ are the corresponding weights, as hyper-parameters controlling the relative importance of the three objectives.

\subsection{Training UAA-GAN}
Training UAA-GAN is an iterative process that $G$ and $D$ perform alternating gradient descent over mini-batches.
In the first step, we sample a batch data from the training set as input images, and send them to to the fixed $G$ to generate fake images.
Both the real images and fake images are then sent to $D$ to calculate the GAN loss for $D$ ($\mathcal L_{GAN\_D}$) for the optimization of $D$'s parameters.
In the next step, we fix the parameters of $D$, and send the generated fake images to $D$ to calculate the GAN loss for $G$ ($\mathcal L_{GAN\_D}$). 
$G$ is then optimized by the weighted sum of three losses (i.e., $\mathcal L_{GAN\_G}$, $\mathcal L_{recon}$ and $\mathcal{L}_{metric}$).
For each batch, $D$ and $G$ are alternately optimized in such training process.

\section{Experiments}\label{experiment}
\subsection{Datasets and Evaluation Protocols}
We evaluate the proposed UAA-GAN on three tasks: image retrieval, person ReID and face recognition. Public benchmark datasets are used in our experiments: Oxford5K\cite{philbin2007object} and Paris6K\cite{philbin2008lost} for image retrieval, Market-1501\cite{zheng2015scalable} and DukeMTMC-ReID \cite{zheng2017unlabeled} for person ReID and FaceScrub\cite{ng2014data} for face search. The set of evaluation datasets include:

\subsubsection*{Evaluation Datasets} 
\begin{itemize}
  \item \textbf{Oxford5K} is the Oxford Buildings Dataset, which contains 5062 images collected from Flickr. It offers a set of 55 queries for 11 landmark buildings, five for each landmark.
  \item \textbf{Paris6K}, similar to Oxford5k, contains 6,412 images that correspond to 12 Paris landmarks with 55 queries.
  \item \textbf{Market1501} is consists of 32,688 annotated bounding boxes of 1,501 individuals. 751 persons' images are used for training and 750 persons' are used for testing.
  \item \textbf{DukeMTMC-ReID} is constructed from a large-scale multi-target multi-camera tracking dataset DukeMTMC.
  We use 702 persons for training and the remaining 702 persons for testing from DukeMTMC-ReID.
  \item \textbf{FaceScrub} contains unconstrained 107,818 face image of 530 celebrities (265 males and 265 females), with around 200 images per person in average. We split the 530 celebrities into 480 for evaluation and 50 for training.
  \end{itemize}
  
  For the training of UAA-GAN, we used additional datasets such as retrieval-SfM-30k \cite{radenovic2016cnn}, ILSVRC  \cite{ILSVRC15}. All datasets used for training include:
  
\subsubsection*{Training Datasets} 
\begin{itemize}
  \item \textbf{retrieval-SfM-30k} \cite{radenovic2016cnn} is composed of 30,012 images, downloaded from Flickr using keywords of landmarks, and from 162 Structure-from-Motion(SfM) 3D reconstruction models. When training UAA-GAN, we only use 1,691 query images in retrieval-SfM-30k.
  \item \textbf{ILSVRC} \cite{ILSVRC15} is the subset of ImageNet containing 1000 categories and 1.2 million images. We randomly selected 2000 images from the validation set of ILSVRC dataset for training UAA-GAN.
  \item \textbf{Market1501} and \textbf{DukeMTMC-ReID}'s training sets are used for training the attack model for ReID.
  \item \textbf{FaceScrub} is used for training UAA-GAN for face search. We randomly selected 1000 face images from FaceScrub, containing 50 persons from FaceScrub's training set with 20 pictures for each person.
  \end{itemize}

\subsubsection*{Evaluation metrics}
Although the image retrieval, person ReID and face search are similar tasks,  the evaluation metrics of these tasks are slightly different.
We follow widely-used evaluation metrics reported in other research papers, 
We use mean Average Precision (mAP) as the evaluation protocol for image retrieval,
mAP and the Cumulative Matching Characteristics (CMC) at rank-1, rank-5 and rank-10 to evaluate ReID,
and CMC at rank-1, rank-5 and rank-10 to evaluate face search.
We evaluate the attack performance of UAA-GAN by comparing the values of these metrics before and after the adversarial attacks.

\subsection{Implementation and Parameter Settings}

All of our models and experiments are implemented in PyTorch framework. 
The experimental server is equipped with 4 NVIDIA TITAN Xp GPUs, 4 Intel Xeon Silver 4110 @ 2.10GHz CPUs and 128GB of RAM.
The weights of the convolutional layers are initialized with a normal distribution with zero mean and standard deviation of 0.02.
We use Adam as the optimization algorithm for discriminator and generator with the following hyper-parameters: $beta1=0.9$, $beta2=0.999$, $epsilon=1e-8$.
The model is trained for 500 epochs and the initial value of the learning rate set to is 0.001 for the generator and 0.004 for the discriminator, in addition the learning rate decays by an additional $10\%$ of its previous value at the 150th epoch and 200th epoch respectively.
For all experiments, The $\epsilon$ in Equation \ref{eq1} is set to 0.1, and the margin $m$ in metric loss is set to 1.
The values of batch size,  $\lambda_m$, and $\lambda_r$ in Equation \ref{eq9} is different for the three tasks.
We set batch size to 32, $\lambda_m = 0.03$, and $\lambda_r = 4$ for image retrieval;
batch size to 256, $\lambda_m = 0.05$, and $\lambda_r = 8$ for person ReID;
batch size to 64, $\lambda_m = 0.01$, and $\lambda_r = 2$ for face search.


\subsection{Adversarial Attacks Results}
In this subsection we show the evaluation results of UAA-GAN on three tasks: image retrieval, person-ReID and face search, respectively.

\begin{figure*}[htbp]
  \centering
  \includegraphics[width=17.8cm]{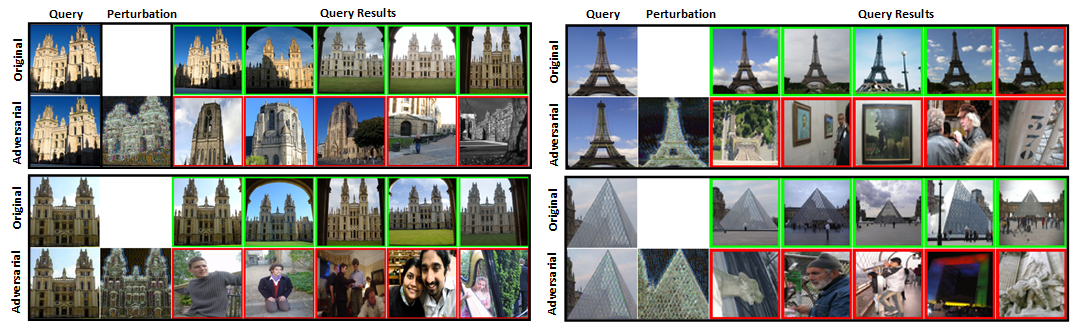}
  \caption{Examples of adversarial attack results of UAA-GAN trained on VGG16 with GeM aggregate function.
  }
  \label{fig3}
\end{figure*}

\begin{table}[htbp]
  \centering
  \caption{Detailed experimental results of adversarial attacks on image retrieval.}
  \label{tabel2}
  \begin{center}
    \begin{tabular}{c|c|c|cc}
          \hline
                                  &                       &           & Oxford5K      & Paris6K    \\
          \hline
          Attacked    &Aggregate      & \multirow{2}{*}{Query}            & \multirow{2}{*}{mAP}           & \multirow{2}{*}{mAP}    \\
          Base Network  &Function &  &  &   \\
          \hline
          \multirow{12}{*}{VGG16}  & \multirow{4}{*}{MAC} &Original       &0.818     &0.788     \\
                                  &                       &Gaussian       &0.794     &0.784     \\
                                  &                       &UAA-SfM        &\textbf{0.080}     &\textbf{0.169}     \\ 
                                  &                       &UAA-ILSVRC     &\textbf{0.044}     &\textbf{0.155}     \\ \cline{2-5} 
                                  & \multirow{4}{*}{RMAC} &Original       &0.748     &0.808     \\
                                  &                       &Gaussian       &0.725     &0.808     \\
                                  &                       &UAA-SfM        &\textbf{0.077}     &\textbf{0.287}     \\ 
                                  &                       &UAA-ILSVRC     &\textbf{0.081}     &\textbf{0.201}     \\ \cline{2-5} 
                                  & \multirow{4}{*}{GeM}  &Original       &0.849     &0.860     \\
                                  &                       &Gaussian       &0.847     &0.863     \\
                                  &                       &UAA-SfM        &\textbf{0.093}     &\textbf{0.177}     \\ 
                                  &                       &UAA-ILSVRC     &\textbf{0.112}     &\textbf{0.260}     \\ \cline{2-5} 
          \hline
      \multirow{12}{*}{ResNet101}  & \multirow{4}{*}{MAC} &Original       &0.769     &0.852     \\
                                  &                       &Gaussian       &0.756     &0.860     \\
                                  &                       &UAA-SfM        &\textbf{0.182}     &\textbf{0.329}     \\ 
                                  &                       &UAA-ILSVRC     &\textbf{0.206}     &\textbf{0.415}     \\ \cline{2-5} 
                                  & \multirow{4}{*}{RMAC} &Original       &0.731     &0.841     \\
                                  &                       &Gaussian       &0.712     &0.840      \\
                                  &                       &UAA-SfM        &\textbf{0.124}     &\textbf{0.352}     \\ 
                                  &                       &UAA-ILSVRC     &\textbf{0.137}     &\textbf{0.326}     \\ \cline{2-5} 
                                  & \multirow{4}{*}{GeM}  &Original       &0.862     &0.907     \\
                                  &                       &Gaussian       &0.763     &0.865     \\
                                  &                       &UAA-SfM        &\textbf{0.137}     &\textbf{0.262}     \\ 
                                  &                       &UAA-ILSVRC     &\textbf{0.159}     &\textbf{0.327}     \\ \cline{2-5} 
          \hline
      \end{tabular}
  \end{center}
\end{table}

\subsubsection{Results of Image Retrieval}

We evaluate the performance of UAA-GAN on two target base networks (VGG16 and ResNet101) with three aggregate functions (MAC, RMAC, GeM). In total, we performed attacks on six target networks. 
All target image retrieval models are fine-tuned on retrieval-SfM-30k \cite{radenovic2016cnn}, using the publicly released code and pre-trained models.
We trained our UAA-GAN on two datasets, one using SfM landmark images (similar to Paris6K and Oxford5k) and the other using more generic images from ImageNet (ILSVRC).
In addition, in order to compare the effectiveness of the generated perturbations, we use the retrieval results of query images with added Gaussian noise of the same noise level as the UAA-GAN generated perturbations.
The results of the attacks are shown in Table \ref{tabel2}.
It is evident that UAA-GAN resulted in a significant drop in retrieval precision on all listed  retrieval models, suggesting that the attacks are effective cross datasets and models. For instance, the original mAP for retrieval are 0.818, 0.748 and 0.849 for MAC, RMAC and Gem on Oxford5K respectively. The figures drop drastically to 0.08/0.044, 0.077/0.081 and 0.093 and 0.112 after UAA-GAN's attack, showing that the retrieval system is completely disrupted by the perturbation. Similar results are present on Paris6K, with the retrieval mAP diminishes by an order of magnitude.
In contrast, the same level of Gaussian noise does not have a significant impact on the retrieval results.
In particular, UAA-GAN trained on unrelated ILSVRC dataset has also achieved remarkable results.
That is to say, UAA-GAN can completely disrupt the image retrieval system by training even on an irrelevant dataset without knowing any domain knowledge of the target networks.
Figure. \ref{fig3} shows several examples of adversarial attack results of VGG16 with GeM aggregate function on both Oxford5K and Paris6K datasets.
We also notice that the generated adversarial query images show little visual difference from the original query image.

\begin{table*}[htbp]
  \centering
  \caption{Detailed experimental results of adversarial attack on person re-identification.}
  \label{tabel3}
  \begin{center}
    \begin{tabular}{c|c|cccc|cccc}
      \hline
                                &           & \multicolumn{4}{c|}{Market1501}       & \multicolumn{4}{c}{DukeMTMC-ReID}     \\ \hline
                                & Query     & mAP     & top1   & top5   & top10    & mAP    & top1    & top5    & top10   \\ \hline
      \multirow{3}{*}{ResNet50} & Original    & 0.722    & 0.895  & 0.955  & 0.971    & 0.650   & 0.809   & 0.907   & 0.934   \\
                                & Gaussian  & 0.537    & 0.736  & 0.855  & 0.920    & 0.421   & 0.662   & 0.789   & 0.831   \\
                                & UAA-GAN   & \textbf{0.010}    & \textbf{0.007}  & \textbf{0.020}  & \textbf{0.030}    & \textbf{0.009}   & \textbf{0.015}   & \textbf{0.035}   & \textbf{0.052}   \\ \hline
      \multirow{3}{*}{MGN}      & Original    & 0.870    & 0.945  & 0.983  & 0.989    & 0.874   & 0.897   & 0.942   & 0.953   \\
                                & Gaussian  & 0.826    & 0.914  & 0.968  & 0.980    & 0.700   & 0.822   & 0.912   & 0.934   \\
                                & UAA-GAN   & \textbf{0.116}    & \textbf{0.129}  & \textbf{0.227}  & \textbf{0.283}    & \textbf{0.045}   & \textbf{0.047}   & \textbf{0.089}   & \textbf{0.114}   \\ \hline

    \end{tabular}
  \end{center}
\end{table*}

\begin{table*}[htbp]
  \centering
  \caption{Transfer results for image retrieval models. 
  The first column represents the target model used to train the UAA-GAN, and the first row indicates the retrieval performance of the original model without attack.
  Each subsequent row represents the retrieval performance of these six search models after being attacked by UAA-GAN.
  All models are evaluated on Paris6K dataset.}
  \label{tabel5}
  \begin{tabular}{c|cccccc}
  \hline
  Perturbations generated from	&	VGG-MAC	&	VGG-RMAC	&	VGG-GeM	&	ResNet-MAC	&	ResNet-RMAC	&	ResNet-GeM	\\ 
  \hline                              
  no attack                     &	0.788 	&	0.808 	&	0.860  	&	0.852 	&	0.841 	&	0.907 	\\  \hline
  VGG-MAC	                      &	0.169 	&	0.260 	&	0.158 	&	0.384 	&	0.407 	&	0.324 	\\ 
  VGG-RMAC	                    &	0.249 	&	0.287 	&	0.204 	&	0.449 	&	0.443 	&	0.395 	\\ 
  VGG-GeM	                      &	0.174 	&	0.252 	&	0.177 	&	0.399 	&	0.398 	&	0.362 	\\ 
  ResNet-MAC	                  &	0.515 	&	0.447 	&	0.449 	&	0.329 	&	0.334 	&	0.273 	\\ 
  ResNet-RMAC	                  &	0.530 	&	0.493 	&	0.484 	&	0.366 	&	0.352 	&	0.299 	\\ 
  ResNet-GeM	                  &	0.491 	&	0.462 	&	0.430 	&	0.325 	&	0.319 	&	0.262 	\\ 
    \hline
  \end{tabular}
\end{table*}

\subsubsection{Results of Person Re-identification}
\begin{figure}[htbp]
  \centering
  \includegraphics[width=8.5cm]{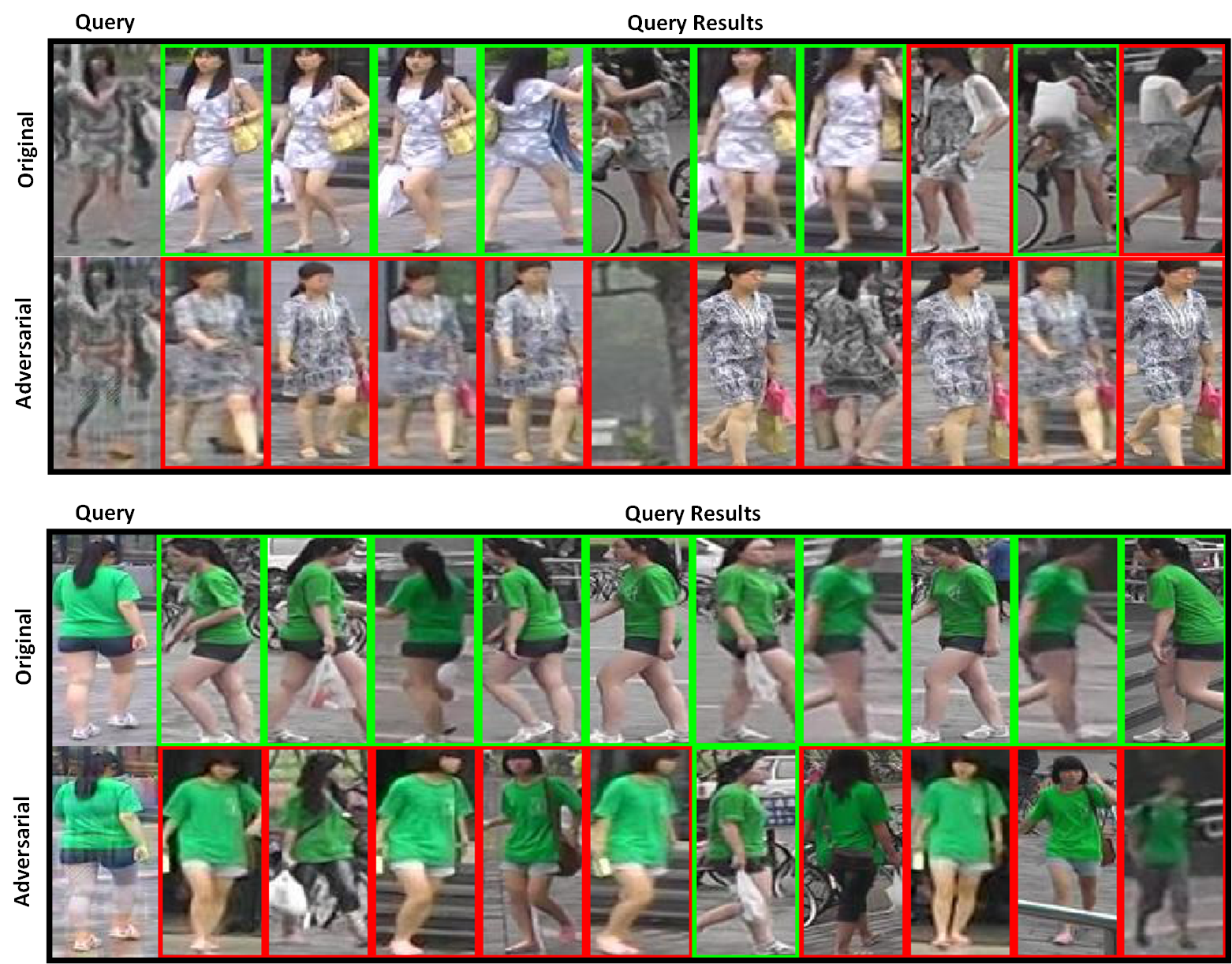}
  \caption{
    Examples of adversarial attack results on Market1501. 
    }
  \label{fig4}
\end{figure}

For person-ReID, we study two typical network architectures, i.e., fine-tuned ResNet50 and MGN \cite{wang2018learning}.
Both networks are trained on DukeMTMC-ReID and Market1501, forming four target networks to attack.
These target networks are trained by the public released codes.
The attack results are summarized in Table \ref{tabel3}.
We observe that, on the four target networks, all the evaluation metrics dropped by a large margin.
The mAP of fine-tuned ResNet50 decreased from 0.722 to 0.010 on Market1501 and from 0.65 to 0.009 on DukeMTMC-ReID respectively.
Meanwhile, the rank-N accuracy has also declined dramatically.
The fact that the same level of Gaussian noise only slightly reduces the mAP of the target models, by around 0.2, further proves the effectiveness of UAA-GAN.
From empirical results, we confirm that both fine-tuned ResNet50 and MGN are very vulnerable to adversarial query images generated by UAA-GAN.
Compared to fine-tuned ResNet50, MGN proves more robust to adversarial attacks.
Under the same parameter settings, the mAP of fine-tuned resnet50 is reduced to less than 0.010, but the mAP MGN model remains 0.116 on Market1501 and 0.045 on DukeMTMC-ReID.
MGN is a part-based methods, which integrates global and local information with various granularities, which could be the reason for its robustness to perturbations.
Some adversarial attack results of MGN on Market1501 are presented in  Figure. \ref{fig4}.

\begin{table}[htbp]
  \centering
  \caption{Detailed experimental results of adversarial attacks on face search.}
  \label{tabel4}
  \begin{center}
    \begin{tabular}{l|l|lll}
      \hline
      \multicolumn{1}{c|}{} & \multicolumn{1}{c|}{} & \multicolumn{3}{c}{FaceScrub} \\ \hline
                                              & Query        & top1 & top5 & top10 \\ \hline
      \multirow{2}{*}{Sphereface}             & Original & 0.713  & 0.809  & 0.850  \\
                                              & UAA-GAN & \textbf{0.215 }  & \textbf{0.355 }  & \textbf{0.424 }  \\ \hline
      \end{tabular}
  \end{center}
\end{table}

\subsubsection{Results of Face Search}
For face search, we tested the attack performance of UAA-GAN on a pre-trained Sphereface model
In Table \ref{tabel4}, we list the query performance of Sphereface on the FaceScrub dataset, and the results after adversarial attacks.
As can be seen from the results, the evaluation suggests that the face search model's accuracy has dropped significantly.
Even the Rank-10 accuracy is below 0.5, which means that we can hardly find the query person from the top 10 results.
The results  demonstrate that the face search model has almost no defence ability against the adversarial attacks from UAA-GAN.


\subsection{Further Analysis}

\subsubsection{Transferability}
The transferability of CNN networks in image classification tasks has been studied in many literatures, it means that the adversarial examples generated for one model will also mislead other models.
We show that in image retrieval, the transferability of adversarial examples also widely exists.
We evaluate the performance of transfer attack by cross-evaluating the generated networks trained for different target networks. 
Quantitative results are summarized in Table. \ref{tabel5}.
Experiments reveal that transferability  exists commonly between different aggregate functions within the same base network structure.
For example, the UAA-GAN trained on VGG-MAC even has a better transfer attack performance on VGG-RMAC and VGG-GeM compared with its original UAA-GAN.
In addition, UAA-GAN also exhibits transferability crossing different network structures, but the transferability becomes weaker.
For example, the UAA-GAN trained on ResNet-MAC only leads slight mAP drop on VGG-based image retrieval model: VGG-MAC(from 0.788 to 0.515), VGG-RMAC (from 0.807 to 0.447), VGG-GeM (from 0.860 to 0.449).
Similar phenomena also occur when using UAA-GAN, which is trained on VGG-based model, to attack the ResNet-based image retrieval models.

\subsubsection{Effect of the Discriminator}
\begin{figure}[htbp]
  \centering
  \includegraphics[width=8.5cm]{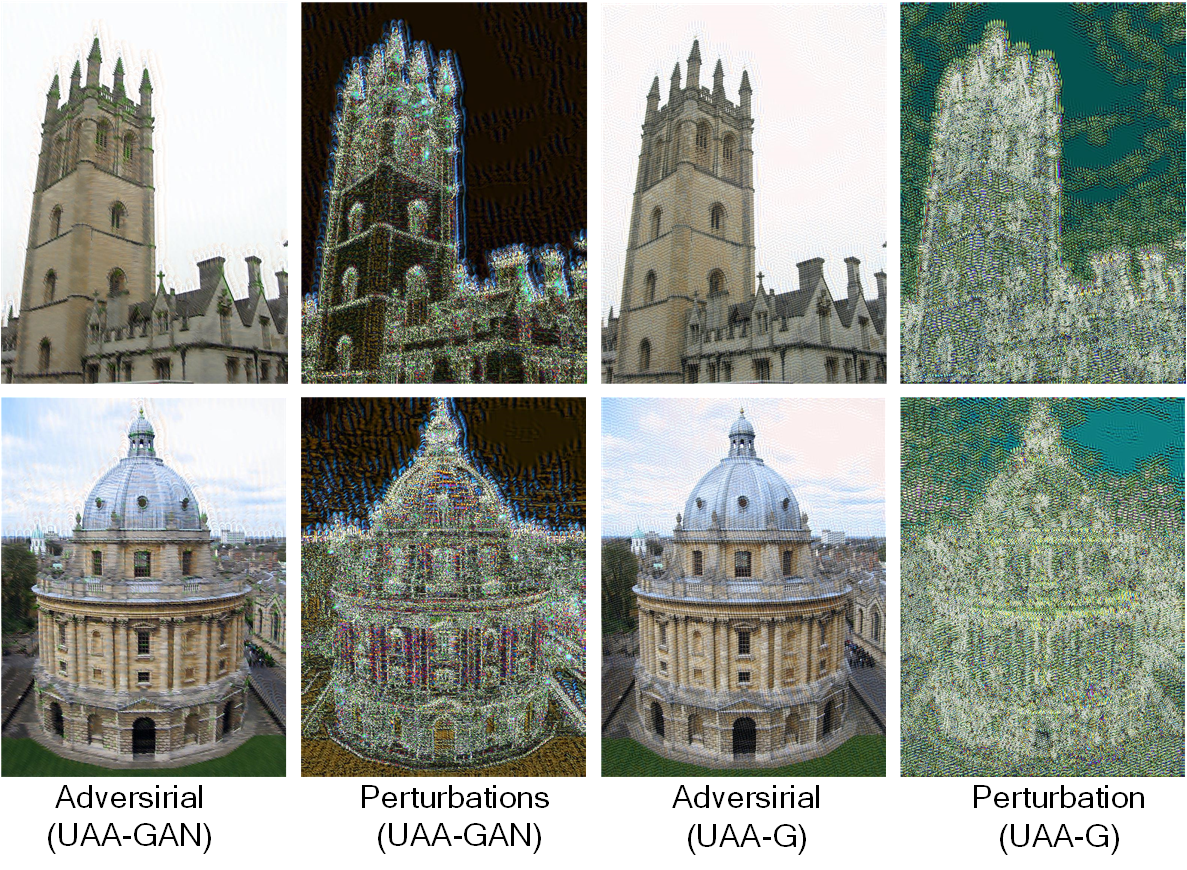}
  \caption{
    Examples of perturbations and adversarial query images generated by UAA-GAN and UAA-G. 
    }
  \label{fig5}
\end{figure}

We further set up an experiment to study the role of the discriminator in the process of training UAA-GAN.
We create a new adversarial attack network without the discriminator of UAA-GAN, which we named UAA-G.
And, keeping all parameters consistent with UAA-GAN, we train the UAA-G on SfM dataset and choose the VGG-GeM as target model.
The results show that UAA-G has similar attack results as UAA-GAN, which achieved an mAP of 0.091 on oxford5k compared to 0.093 of UAA-GAN.
However, the perturbations generated by UAA-G are more abrupt than UAA-GAN's, and it also results in lower quality of the generated adversarial examples.
The details of the qualitative comparison of the generated perturbations are shown in Figure. \ref{fig5}.
Through the comparison, we can draw the conclusion that the discriminator will force the generator to produce more image-realistic and less-distinguishable adversarial examples, by incorporating subtle perturbations in textured or salient areas of the images, such as key body parts of human, dominant structural patterns/textures or edges, rather than in visually insignificant areas.

\section{Conclusion}\label{conclusion}
In this paper, we investigate the effect of adversarial examples for deep feature-based image retrieval.
We proposed UAA-GAN model, which can effectively generate visually imperceptible adversarial query image to attack the image retrieval systems, making them return content irrelevant results.
UAA-GAN is an unsupervised learning algorithm, which only requires a small amount of unlabeled images for training.
Once trained, it is able to produce query-specific perturbation for any input image.
We conducted extensive experiments on three tasks and five datasets, illustrating that UAA-GAN is able to significantly cripple the performance of retrieval systems without obtrusive changes in the query images.
Through visual comparison, we demonstrate that the discriminator enforces the UAA-GAN to produce less perceptible perturbations. 
We also show that the adversarial examples generated by UAA-GAN have transferrable attack abilities across different target models.

\bibliographystyle{IEEEtran}
\bibliography{IEEEabrv,IEEEexample}

\end{document}